\documentclass{bmvc2k}

\usepackage{times}
\usepackage{epsfig}
\usepackage{graphicx}
\usepackage{amsmath}
\usepackage{amssymb}
\usepackage{placeins}
\usepackage{multirow}
\usepackage{colortbl}
\usepackage{xcolor}
\usepackage{color,soul}
\usepackage{ulem}
\usepackage{xcolor}

\definecolor{mybluelight}{rgb}{0.9, 0.9, 1.}
\definecolor{mybluelight2}{rgb}{0.64453125, 0.78125, 0.87890625}
\definecolor{myorangelight2}{rgb}{0.99609375, 0.79296875, 0.6171875}

\title{Holistic Guidance for Occluded Person Re-Identification}

\addauthor{Madhu Kiran}{madhu_sajc@hotmail.com}{1}
\addauthor{R Gnana Praveen}{praveengnan.24@gmail.com}{1}
\addauthor{Le Thanh Nguyen-Meidine}{le-thanh.nguyen-meidine.1@ens.etsmtl.ca}{1}
\addauthor{Soufiane Belharbi}{soufiane.belharbi.1@ens.etsmtl.ca}{1}
\addauthor{Louis-Antoine Blais-Morin}{lablaismorin@GENETEC.COM}{2}
\addauthor{Eric Granger}{Eric.Granger@etsmtl.ca}{1}
\addinstitution{
LIVIA, Dept. of Systems Engineering\\
École de technologie supérieure\\
Montreal, Canada
}
\addinstitution{
 Genetec Inc.\\
 Montreal, Canada
}

\runninghead{Kiran ET AL.}{Holistic Guidance for Occluded Person Re-Identification}

\def\ie{\emph{i.e}\bmvaOneDot}

\begin{document}

\maketitle

\begin{abstract}
In real-world video surveillance applications, person re-identification (ReID) suffers from the effects of occlusions and detection errors. Despite recent advances, occlusions continue to corrupt the features extracted by state-of-art CNN backbones and thereby deteriorate the accuracy of ReID systems. To address this issue, methods in the literature rely on an additional costly process, such as pose estimation, where pose maps provide supervision to focus on visible parts of occluded regions. 
In contrast, we introduce a Holistic Guidance (HG) method that relies on holistic (or non-occluded) data and its distribution in the dissimilarity space to train the CNN backbone on an occluded dataset. This method is motivated by our empirical study, where the distribution of pairwise between-class and within-class matching distances (Distribution of Class Distances or DCDs) between images has considerable overlap in occluded datasets compared to holistic datasets. Hence, our HG method employs this discrepancy in DCDs of both datasets for joint learning of a student-teacher model to produce an attention map that focuses primarily on visible regions of the occluded images. In particular, features extracted from both datasets are jointly learned using the student model to produce an attention map that allows dissociating visible regions from occluded ones. Additionally, a joint generative-discriminative CNN backbone is trained using a denoising autoencoder such that the system can self-recover from occlusions. 
Extensive experiments on several challenging public datasets indicate that the proposed approach can outperform state-of-the-art methods on both occluded and holistic datasets. \textbf{Our code is available}\footnote{\url{https://github.com /madhukiranets/HolisitcGuidanceOccReID2}}.
\vspace{-0.1cm}
\end{abstract}


\begin{figure}
    \centering
    \includegraphics[width=0.9\linewidth]{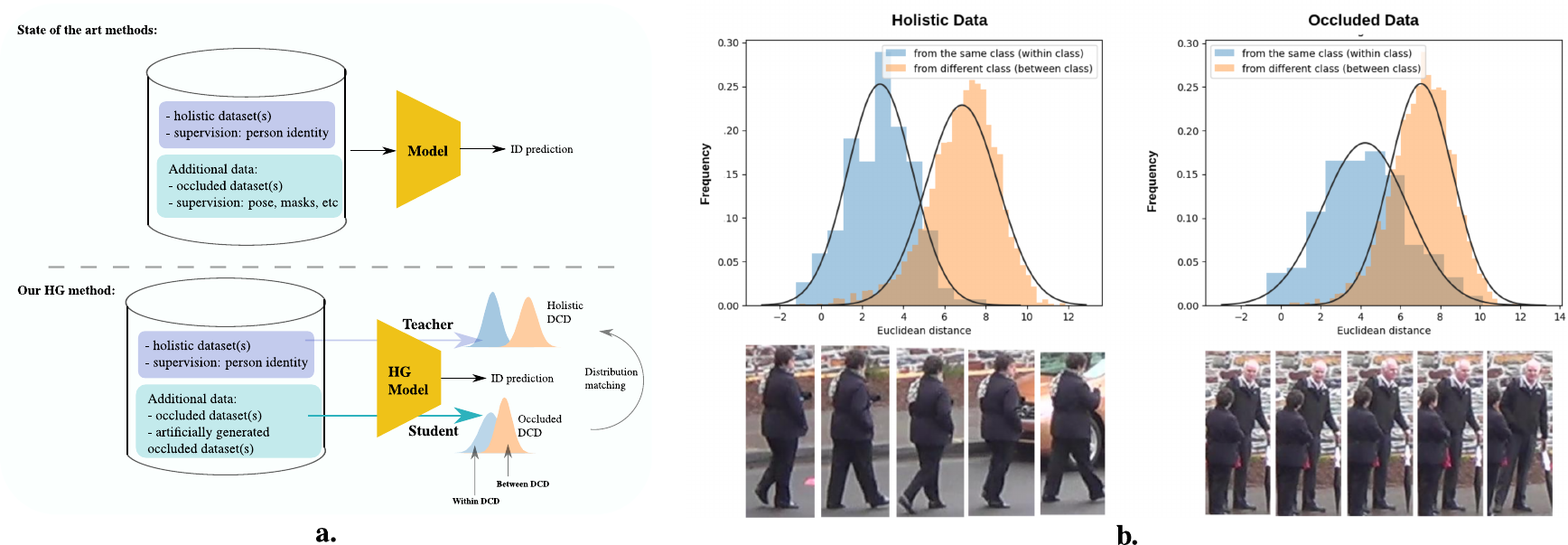}
    \caption{
    \textbf{(a)} An illustration of approaches to address occlusion in person ReID during training. 
    \textbf{Top}: State-of-the-art models require additional supervision and occluded datasets. \textbf{Bottom}: Our proposed HG method requires no additional supervision but relies only on an additional holistic dataset for reference to non-corrupted features. 
    \textbf{(b)} Examples of Class Distance Distributions of Duke-MTMC (\textbf{left}) and Occluded-Duke-MTMC (\textbf{right}) datasets measured in the distance space. The \colorbox{mybluelight2}{\textbf{blue DCD}} shows the within-class distribution, while the \colorbox{myorangelight2}{\textbf{orange DCD}} shows the between-class distribution. For Occluded-Duke-MTMC, within-class distances are relatively high and overlap with between-class distances.
    }
    \label{fig:method}
    \vspace{-0.2cm}
\end{figure}

\section{Introduction}



Person Re-Identification (ReID) systems seek to associate individuals captured over a distributed set of non-overlapping camera viewpoints. This key visual recognition task has recently drawn significant attention due to its wide range of applications, e.g., autonomous driving, pedestrian tracking, sports analytics, and video surveillance~\cite{farenzena2010person,panda2017unsupervised,dukeDUPLICATED,zhang2016learning}. Despite the recent progress with deep learning (DL), person ReID remains a challenging task in real-world applications due to the non-rigid structure of the human body, variability of capture conditions (e.g., illumination, scale, motion blur), in addition to person detection issues like miss-alignment, background clutter, and occlusion~\cite{teacherstudent,partial2,occ_pgfa}.  

This paper focuses on the occlusion issue for person ReID, a challenge that has attracted much attention~\cite{gao2020pose,partial1,partial2,occ_pgfa,occ_mhsa,higherorder}. When bounding boxes or regions of interest (ROI) are occluded, the CNN backbone extracts noisy features, leading to pairwise matching errors between query and reference ROIs and poor ReID accuracy for the occluded class. 
Since occlusions are diverse in color, shape, and size, extracting features from the entire ROI can potentially corrupt the global representation. 

Several authors have attempted to address occluded person ReID by using pedestrian detectors that can additionally refine person ROIs~\cite{occ_rcnn}.  Other methods follow an intuitive solution of masking occluded regions, extracting occlusion aware features, or applying weights and masks to occluded regions, applying body masks based on pose estimation,  etc. ~\cite{gao2020pose,occ_pgfa, occ_mask}. With these state-of-the-art methods, the external mechanism for mask generation add a considerable time complexity during run-time. Unlike these methods, we propose a new method that only requires person identity labels as supervision and does not rely on additional supervision such as pose estimation. This provides robustness to occlusion in person ReID, yet lower complexity during inference.

Our method is motivated by the fact that the distribution of features learned from occluded and holistic\footnote{In the person ReID, the term "holistic" refers to image data that contains the full body of a person in both query and gallery sets. Holistic (or non-occluded) datasets have fewer occluded samples w.r.t. the overall dataset size.} ReID datasets are different. The dissimilarity space can be regarded as space defined by dissimilarity coordinates, and CNN features are transformed into that space by computing pairwise matching distances for within-class and between-class samples in a given batch. It has been shown to successfully learn to separate feature representations for data that is noisy and overlapping~\cite{dissim_main, dissim_main2}.  Occluded person ReID is a good example of a problem with class overlap. Inspired by~\cite{costa2020dissimilarity, dissimilar2,dissimilar1,dissim_main}, we consider the dissimilarity space to capture the discrepancy between images in occluded and holistic datasets. 

\begin{figure}
 \centering
\includegraphics[width=0.6\linewidth]{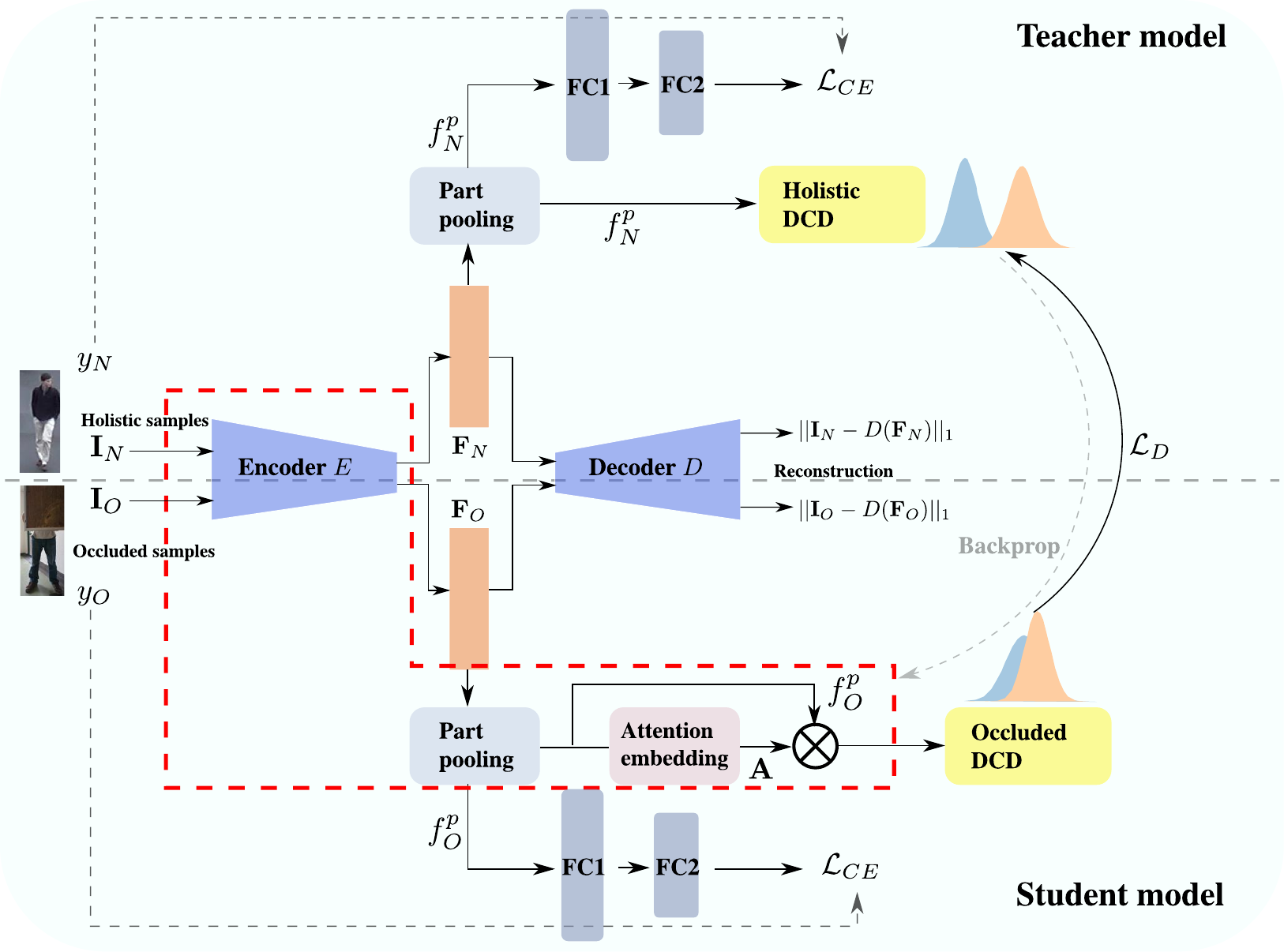}
   \caption{Our proposed HG method where a teacher model uses a holistic data distance distribution to train the student network(trained on artificially occluded or real occluded samples) such that it can accurately recognize persons appearing in occluded images. 
   }
  \label{fig:vae} 
  \vspace{-0.2cm}
\end{figure}

Fig.~\ref{fig:method} shows the distribution of within-class and between-class distances (DCDs) among pairs of samples extracted from occluded and holistic datasets. We note two aspects:
\textbf{(1) Within-class DCDs}: In an occluded dataset, this distance tends to be greater, with high variance, compared to holistic or non-occluded cases. Such distances are normally expected to be lower due to the similarity among samples within the same class. 
\textbf{(2) Overlap within- and between-class DCDs}: Large within-class DCDs are caused by samples being pushed away from the same class, allowing for substantial overlap with samples from other classes. The overlap between samples of different classes is more likely to impede discrimination among classes, leading to poor recognition. Note that both datasets used to generate Fig.\ref{fig:method} -- Duke-MTMC (Holistic)~\cite{dukeDUPLICATED} and Occluded-Duke~\cite{occ_pgfa} -- are from the same domain, but occlusions still cause such discrepancies. Based on these observations of DCDs, and considering the poor performance of models over occluded datasets, we hypothesize that occlusion is a potential source of corruption for feature representations in person ReID tasks. 

This paper proposes a Holistic Guidance (HG)  student-teacher network that relies on the distribution of holistic data in the dissimilarity space to train a student (CNN backbone) on an occluded dataset. The discrepancy of within- and between-class DCDs across datasets allows the network to extract features on occluded samples while simultaneously maintaining a good between-and within-class separation. Models trained on occluded data tend to overfit due to class overlap, so we advocate for using guidance from non-corrupted features of a larger holistic data in the dissimilarity space to mitigate this issue.  Although both datasets can have different identities, transforming the samples to the dissimilarity space translates into a  binary classification problem~\cite{dissim_main,costa2020dissimilarity}. In practice, such learning scenarios could be achieved by using a single holistic dataset, and by building an artificially (augmented) occluded dataset. A second alternative involves using a training set consisting of real occluded samples. Our method performs HG since it relies on features with properties learned from holistic data to guide the CNN backbone in learning features of an occluded dataset.

Our method (shown in Fig.\ref{fig:vae}) is also comprised of a shared generative model (i.e., a denoising auto-encoder) that is trained simultaneously on both datasets to enable self-recovery from occlusion. The student model has an additional embedding for producing an attention map, allowing the partial or local features to re-focus attention only on visible body-part features while ignoring the occluded regions that cause distribution discrepancy. Several authors have proposed generative models for person ReID \cite{occlusionrecovery,jointgen}, mainly for GAN based data augmentation. In contrast, we introduce a denoising auto-encoder as the CNN backbone for our HG student-teacher network, allowing to self-recover in cases of occlusion.

\noindent \textbf{Main contributions:}
(1) A novel HG student-teacher network that relies on the distribution of holistic data in the dissimilarity space to train a CNN backbone on the occluded dataset.
(2) To motivate our HG method, we show that within-class DCDs of Occluded-ReID datasets overlap with between-class DCDs by a larger margin than holistic ReID datasets, even in cases where the occluded dataset is a subset of a holistic one.
(3) Extensive experiments were performed on challenging Occluded~\cite{icmeocc, occ_pgfa}, Partial~\cite{partialilids,partialdata}, and Holistic~\cite{zheng2015scalable,dukeDUPLICATED} ReID datasets show that our HG method can outperform many SOTA methods.

\vspace{-0.4cm}
\section{Related Work in Person-ReID}
\vspace{-0.2cm}

\noindent \textbf{Image-Based Methods:} 
Siamese Networks have first been used in~\cite{yi2014deep} that employs three Siamese sub-networks for deep feature learning. Most of the further work based on  deep-architecture 
ReID~\cite{ahmed2015improved,cheng2016person,chen2017beyond,liu2017end,varior2016gated} approaches introduce an end-to-end ReID framework, where both feature embedding and metric learning 
have been investigated. A few attention-based approaches for deep ReID ~\cite{li2017learning,su2017pose,zhao2017deeply} address misalignment challenges by incorporating a regional attention sub-network into a base re-ID model. 
With part-based methods, 
local features are extracted from different regions to enhance the discriminative power of the features. Suh et al.~\cite{suh2018part} extracted parts from the feature map and trained each part with separate classifiers.   ~\cite{humanparsing} used parts method to extract local features. In addition to this, other methods such as  ~\cite{sun2018beyond,zhao2017deeply} have used part pooling with attention for a refined partial feature.

\noindent \textbf{Occluded Person ReID:} 
Occluded person ReID is different from person ReID or holistic person ReID because during test time, the probe images are often occluded as in real-world applications. Hence ~\cite{icmeocc} have proposed to use a binary classifier to classify the images as occluded or not to distinguish occluded ones from holistic images.~\cite{occ_pgfa} considered using pose guided feature alignment to align part features or local features. Similar to ~\cite{occ_pgfa}, other works align local or part features by pose estimation like ~\cite{poseinvariant,globallocal}. Similar to this, ~\cite{higherorder} use pose estimator to help predict key-points on body parts and use graph-based methods. The main disadvantage of pose-aligned methods is that they suffer from requiring additional supervision like the pose estimation step, which is error prone~\cite{alignedReID}. Recently, ~\cite{gao2020pose} use a method where, in the first step, they obtain discriminative features by using pose guided attention and then mine for visible parts. \cite{partial2} have used foreground-background mask instead of the pose. Recently, ~\cite{diverse} used only identity information to learn occluded ReID with part aware transformers. Note that they add additional complexity with the transformers used for attention.~\cite{aaaiOccluded} is an other approach using only identity information, where Occluded ReID is considered as a set matching task to make invariant to occlusion of different parts.

Similar to our method, Zhuo et al.~\cite{teacherstudent} employ a student-teacher model. In the first stage, their teacher network learns from a holistic dataset, where both identification and supervised saliency is learned using a salient object detector. However, it differs from our approach in that: \textbf{(1)} it does not learn any supervised salient region detection, and \textbf{(2)} it allows recovering from small occlusions by using a denoising auto-encoder as a CNN backbone.

\vspace*{-0.55cm}

\section{Proposed Approach}
\vspace{-0.3cm}
 
A Holistic Guidance (HG) method is introduced for unsupervised learning of attention maps in Occluded-ReID, without the need for any external guidance such as pose or segmentation maps. A student-teacher network is proposed, where the teacher network relies on holistic images or non-occluded images to teach the student network the DCD of holistic features. This allows the student to learn an attention map such that, when applied to occluded images, results in uncorrupted features like those of the teacher. More specifically, a joint generative and discriminative backbone is trained with a denoising autoencoder for the student-teacher model. It simultaneously learns to match image pairs while reconstructing images. 

\noindent \textbf{Problem Formulation:} 
Let $I_O$ and $I_{N}$ denote input images from the occluded and holistic dataset, respectively. $N$ and $O$ denote the components from teacher and student trained from holistic data and occluded data, respectively. Let $y_O$ and $y_N$ denote the identity labels of occluded and holistic datasets. $\mathbf{F}_O$ and $\mathbf{F}_N$ represent the global feature maps of occluded and holistic datasets, which are obtained from the shared encoder $E$. The local or part-based features $f_{O}^{i}, f_{N}^{i}$ are produced by applying a pooling function on $\mathbf{F}_O$ and $\mathbf{F}_N$. By minimising the discrepancy between-class distance distribution of  $f_{O}^{i}, f_{N}^{i}$, we intend to learn an attention map $A^{i}$ that is applied on the $f_{O}^{i}$ to obtain features ${f_a}^{i}$ for $i=1,..,p$ for each part. During testing, $\mathbf{\psi}$, which is a concatenation of global and local features, allows extracting features for matching from the gallery and query using a distance function, and to retrieve the identity.

\noindent \textbf{Robust Backbone Model:}
We propose a joint learning framework of a denoising auto-encoder along with the classification networks to be robust to occlusions for person ReID. The input images are augmented by adding small noise using random erasing data augmentation, while the reconstruction loss is obtained using actual images. The encoder $E$ is shared between denoising auto-encoder and ReID classification layers. In order to obtain robust features with both Generative and Discriminative properties, we reconstruct the input images using a Decoder on the embedding  $\mathbf{F}_N$ and $\mathbf{F}_O$. Hence $E$ and $D$ together form a denoising auto-encoder. We have used a denoising autoencoder in order to exploit the full potential of the generative capability to take full advantage of the class distributions of the holistic data-set. Let $\mathbf{\widehat{I_c}}$ represent the reconstructed image, and $\mathbf{F_c}= E(\mathbf{I_c})$ be the latent feature representation of the encoder, 
where $c\in \{N, O\}$ (holistic and occluded images) and $\mathbf{I_c}$ is the input image. The size of $\mathbf{F_c}$ is $B \times C \times w \times h$, where $B$ is the batch, $C$ is the number of output channels of the encoder $E$ and $w,h$ width and height of the feature map. $\mathbf{F_c}$ can also be referred to as the latent feature representation of the auto-encoder. A part-based pooling is applied on $\mathbf{F_c}$ to obtain $p$ parts of stripes of features. Our part pooling method used is similar to~\cite{sun2018beyond} where a 2D feature map is split into horizontal stripes of $p$ parts. Global average pooling is then performed on each of the $p$ parts to obtain $p$ feature vectors of size $C$. Each feature vector of $p$ parts is assigned to a unique classifier, resulting in $p$ classifiers trained using identity labels of the corresponding datasets. 

The predicted output for each given image $\mathbf{I_c}$ from the classifier is $\hat{y}_{i,c}, \text { where } i=1, \ldots, p$ parts. The identity prediction loss function over all the parts is:
\begin{equation}
\label{eqn:xen}
       \mathcal{L}_{\text {CE,c}}=\frac{1}{K} \sum_{i}^{K}-\log \left(\frac{\exp \left(\mathbf{W}_{y_{i}}^{T} \mathbf{x}_{i}+b_{y_{i}}\right)}{\sum_{j=1}^{N} \exp \left(\mathbf{W}_{j}^{T} \mathbf{x}_{i}+b_{j}\right)}\right)
\end{equation}
\noindent where $\mathcal{L}_{\text {CE,c}}$ is the cross entropy loss, $K$ is the batch size, class label $y_{i} \in\{1,2, \ldots, N\}$ is associated with $i^{th}$ training image. $\mathbf{W}_{j}$ and $b_{j}$ are the weights and bias of last fully connected layer for class $y$. Similarly $\mathbf{W}_{j}$ and $b_{j}$ are the weights and bias of the the $j^{th}$ class. The denoising auto-encoder is learned using reconstruction loss between the reconstructed and original images before random erasing augmentation~\cite{randomerasing}, which acts like noise added in denoising auto-encoders. We propose this inspired by~\cite{faceocclusion} to self recover from occlusion. Normally in~\cite{faceocclusion} an autoencoder is learnt to generate non occluded face image from artificially occluded face image which is then post processed for recognition. But in our case since we use a joint representation we expect the joint feature to have the self recovering properties. The reconstruction loss of the autoencoder is given by, 
\begin{equation}
\label{eqn:recons}
       \mathcal{L}_{\mbox{recon,c}}=\mathbb{E}\left[\left\|\mathbf{I_c}-D\left(\mathbf{F_c}\right)\right\|_{1}\right],
\end{equation}
\noindent where $D$ denotes the decoder function of the denoising auto-encoder. We do not use the reconstructed image for any other steps. However, the reconstruction loss is optimised in order to enable the deep features to have generative properties and to self-recover from occlusion. The total loss for the joint learning of generative discriminative learning is:
\begin{equation}
\label{eqn:joint}
       \mathcal{L}_{\mbox{joint},c} =  \mathcal{L}_{\text {CE,c}} + \lambda \mathcal{L}_{\mbox{recon},c} \;,
\end{equation}
\noindent where $\lambda$ is the trade-off parameter.

\noindent \textbf{Student-Teacher Model:}
A student-teacher model with the proposed backbone is shared between the teacher and student. In addition to the backbone, the student model carries an embedding to produce attention maps. From Fig.~\ref{fig:method}, it can be observed that a DCD obtained by comparing extracted deep features of occluded in-class images overlap with those of out-of-class distances by a large margin in comparison with the holistic dataset.  The overlap of the DCD indicates corrupted features as in-class distance distribution must have good separation from that of out-of-class distribution. We further justify the use of holistic guidance by the following. One could learn a separation between classes in the feature space using a triplet or contrastive loss on an occluded dataset alone. Yet, the model may overfit on the occluded dataset due to class overlap. However, holistic datasets are much larger than occluded ones. Therefore, they can provide a good generalization of non-corrupted DCD given that class overlap can be well dealt within the dissimilarity space. By matching the DCD, the student network can learn the class overlap of the teacher network (which is well separated).

Fig.~\ref{fig:vae} shows our overall architecture along with the backbone auto-encoder-based deep feature extractor. We simultaneously take two input images—one from the holistic dataset and the other from the occluded dataset. The extracted deep features are simultaneously optimized for identity loss by learning a set of two fully connected layers for classification.
We use two separate classifiers, one for the teacher to learn holistic data identities and the other for the student to learn occluded data identities.

\emph{Attention Embedding} is particular to student network alone capable of producing attention for the partial features such that the attended partial feature will have good separation between within and between-class distances similar to that of the teacher. Let the attention embedding (which is a set of two layers $1\times1$ convolutional filters with ReLU between them followed by batch normalization layer with sigmoid activation for final attention output) be represented by $AE$. The attention produced by the attention embedding is given by $A^i =  AE(f^i_O)$. 
The attention maps in $A^i$ of size $B \times C$ are obtained for each partial feature, with $i=1, \ldots, p$ and $p$ being the number of parts. The attention  obtained is multiplied  with each partial feature to obtain attended partial features, ${{f_a}^i =  f^i_O  \bigotimes  A^i}$.
The layers for occluded image classification, $FC^i_c$, are applied on each of the attended partial feature ${f_a}^i$. While training on an artificially occluded dataset alone, we use a binary classifier to learn occluded or non-occluded images on artificially occluded samples similar to~\cite{icmeocc}.

In order to \emph{learn the attention}, the student network relies on occluded input images and distance distribution matching. Then, DCD of the occluded and holistic features are compared. Given a mini-batch of image input with occluded and holistic images $I_O$ and $I_N$, partial features $f^i_N$ and ${f_a}^i$ (partial feature with attention) are extracted. We denote the class identity for the features by $u$ and $v$. Therefore, for each mini-batch, we extract pairs of image features with different combinations within a batch according to:
\begin{equation}
\label{eqn:wc}
d_{i}^{\mathrm{wc}}\left(\mathbf{I^{u}_c}, \mathbf{I^{v}_c}\right)=\left\|P^{i,u}_N - P^{i,v}_N\right\|_{2}, u = v \;,
 \ \ \mbox{and,} \ \  
d_{i}^{\mathrm{bc}}\left(\mathbf{I^{u}_c}, \mathbf{I^{v}_c}\right)=\left\|P^{i,u}_N - P^{i,v}_N\right\|_{2}, u \neq v \; .
\end{equation}
Eqn.~\ref{eqn:wc} transforms the features to \textit{dissimilarity space}. $P^i$ denotes the part features \ie, $f^i_N$ for holistic data and ${f_a}^i$ for occluded data. The distance distributions are extracted from $d_{i}^{\mathrm{wc}}$ and $d_{i}^{\mathrm{bc}}$ for both holistic and occluded data. We implicitly learn to produce a good attention embedding $AE$ by minimising the discrepancy between DCD of holistic and occluded data using MMD~\cite{mmd}. Let $\mathbf{D_c^{wc}}$ and $\mathbf{D_c^{bc}}$ be the distributions from $d_{i}^{\mathrm{wc}}$ and $d_{i}^{\mathrm{bc}}$. 
The losses measuring the discrepancy between class distributions of holistic and occluded data are,
\begin{equation}
\label{eqn:wcmmd}
\mathcal{L}_{D}^{\mathrm{wc}}=M M D\left(\mathbf{D}_{N}^{\mathrm{wc}}, \mathbf{D}_{O}^{\mathrm{wc}}\right)  \ \ \mbox{,} \ \  
\mathcal{L}_{D}^{\mathrm{bc}}=M M D\left(\mathbf{D}_{N}^{\mathrm{bc}}, \mathbf{D}_{O}^{\mathrm{bc}}\right)
\ \ \mbox{and,} \ \ 
\mathcal{L}_{\mbox{global}}=M M D\left(f_{N}, f_{a}\right)
\end{equation}

It is important to note that losses $\mathcal{L}_{D}^{\mathrm{wc}}$ and $\mathcal{L}_{D}^{\mathrm{bc}}$ are optimized by fixing  $\mathbf{D_N^{bc}}$ and $\mathbf{D_N^{wc}}$, the DCD of teacher network. This allows the student network distance distribution to match that of the teacher network. Hence, Eqn.~\ref{eqn:wcmmd} calculates the discrepancies between the within-class and between-class DCDs of holistic datasets and occluded datasets. This loss is minimized during learning to obtain a good attention map from the embedding to focus on non-occluded regions of occluded images. $\mathcal{L}_{global}$ calculates the MMD distance between teacher features and student features to encourage the model to perform well on both occluded and holistic data. This $\mathcal{L}_{global}$ is particularly used when the holistic and occluded datasets are from different domains. Parameters $\lambda_1$,  $\lambda_2$, and $\lambda_3$ balance these losses, and are determined empirically. 
\begin{equation}
\label{eqn:lossd}
\mathcal{L}_{D} = \lambda_1 \mathcal{L}_{D}^{\mathrm{wc}}+ \lambda_2 \mathcal{L}_{D}^{\mathrm{bc}} +  \lambda_3 \mathcal{L}_{\mbox{global}} \; .
\end{equation}

\noindent \textbf{End-to-End Learning and Testing:}
The full system is optimized for reconstruction losses and identity losses with cross-entropy on both occluded and holistic data, and finally, the class distance distribution loss. The overall loss function $\mathcal{L}_{Total}$ is given by,
\begin{equation}
\label{eqn:losstotal}
\mathcal{L}_{\mbox{Total}} =  \alpha \mathcal{L}_{\mbox{joint},c} + (1-\alpha) \mathcal{L}_{D} \; ,
\end{equation}
where $\alpha$ balances these losses, and is determined empirically (see  supp. material). During testing, only the student is used along with the components denoted within Fig.~\ref{fig:vae} to extract  $\psi$, which is a concatenation of global and local features, $F_O$ and ${f_a}^i$, for both gallery and query images. Extracted features are matched with Euclidean or Cosine distance to retrieve the identity of a query image from the gallery.

\vspace*{-0.1cm}
\section{Experimental Results and Discussion}


\noindent \textbf{Datasets:}
Our approach is validated on three challenging groups of datasets -- Holistic ReID, Occluded-ReID, and Partial ReID datasets. Our main objective is to address performance on Occluded-ReID problems (Occluded-DukeMTMC~\cite{occ_pgfa} and Occluded-ReID~\cite{icmeocc} datasets) and Partial ReID problems (Partial-ILIDS~\cite{partialilids,occ_pgfa} and Partial-ReID~\cite{partialdata} datasets), but we also evaluated on Holistic problems -- Market1501~\cite{zheng2015scalable} and Duke-MTMC~\cite{dukeDUPLICATED} datasets -- to further assess the effectiveness our approach on regular ReID problems. 
The Occluded-DukeMTMC dataset~\cite{occ_pgfa} contains a total of 15,618 training and 17,661 gallery with 2,210 occluded query images. This is a subset of the Duke-MTMC dataset. To test on Occluded-Duke, we train the student model with Occluded-Duke training set similar to~\cite{occ_pgfa,depthocc,higherorder}. The Occluded-ReID dataset~\cite{icmeocc} mimic real-world application scenarios by collecting datasets using mobile camera equipment on campus. It has a total of 2,000 annotated images with 200 identities. Each identity consists of 5 full-body images and 5 partial images.  The Holistic ReID and Partial ReID datasets are described in the supplementary material.


\noindent \textbf{Implementation Details}: 
For validation, the ResNet50~\cite{he2016deep} was implemented as our backbone Encoder. Transposed convolution layers were used along with Interpolation for the Decoder (see details in the supplementary material). To evaluate Occluded-Duke-MTMC, we train the student using the train data of Occluded-Duke-MTMC, and the teacher with Market1501~\cite{zheng2015scalable}, as in the SOA. Since Occluded-ReID and Partial ReID do not have a prescribed set of training images, the whole dataset was used for testing (as in~\cite{gao2020pose}).  To have a common setting with the SOA~\cite{occ_pgfa, occ_pgfa}, we use an input image size of 384 X 128. We train our backbone with Partial Features with $p=6$ parts, and set the co-efficient $\lambda=0.01$. The teacher network is pre-trained for 15 epochs, and the student-teacher is trained together for 120 epochs.  The Adam optimization is used with an initial learning rate of 0.0003.  We report the Cumulative Matching Characteristics (CMC) and mean average precision (mAP)~\cite{zheng2015scalable}.
\begin{table}[!b]
\vspace{-0.4cm}
\centering
\resizebox{.8\linewidth}{!}{
\begin{tabular}{|l|l|l||rrr|r|}
\hline 
\textbf{Method}  &  \textbf{Backbone}   &  \textbf{Supervision}  & \multicolumn{4}{c|}{\textbf{Accuracy}} \\ 
                    &               &         & Rank-1  & Rank-5 & Rank-10 & mAP               \\\hline\hline  
LOMO+XQDA~\cite{liao2015person}, CVPR 2015  & - & None            & 8.1 & 17.0 & 22.0 & 5.0                             \\
Part Aligned~\cite{zhao2017deeply}, ICCV 2017 & GoogLeNet               & None             & 28.8     & 44.6     & 51.0       & 20.2              \\ 
Random Erasing~\cite{randomerasing}, AAAI 2020 &ResNet50          & None            & 40.5     & 59.6     & 66.8     & 30.0               \\ 
HACNN~\cite{li2018harmonious}, CVPR 2018 & Custom                           & None             & 34.4     & 51.9     & 59.4     & 26.0                \\ 
Adver Occluded~\cite{advoccluded}, CVPR 2018 & ResNet50             & None            & 44.5     & -        & -        & 32.3                 \\ 
PCB~\cite{sun2018beyond}, ECCV 2018 & ResNet50                     & None              & 42.6     & 57.1     & 62.9     & 33.7                 \\ \hline
Part Bilinear~\cite{suh2018part}, ECCV 2018 & GoogLeNet                    &  Occluded-Duke     & 36.9     & -        &          & -                       \\ 
PGFA~\cite{occ_pgfa},ICCV, 2019 & ResNet50                           & Occluded-Duke + PM    & 51.4     & 68.6     & 74.9     & 37.3                  \\ 
Depth Occln~\cite{depthocc}, PRL 2020& ResNet50                   &  Occluded-Duke + PM  & 53.0     & 67.0     & 72.9     & 38.1                    \\ 
HOReID~\cite{higherorder}, CVPR 2020 & ResNet50                           &  Occluded-Duke + KP & 55.1     & -        & -        & 43.8                      \\  
PAT~\cite{diverse}, CVPR 2021 &  ResNet50        &  Occluded-Duke        & 64.5     & -        & -        & 53.6                  \\
MOS~\cite{aaaiOccluded}, AAAI 2021 &ResNet50             &  Occluded-Duke      & 61.0     & -        & -        & 49.2                    \\  

MOS~\cite{aaaiOccluded}, AAAI 2021 &  ResNet50-IBN        &  Occluded-Duke        & \textbf{66.6}     & -        & -        & \textbf{55.1}                  \\ \hline 
HG(ours) &  ResNet50     & Occluded-Duke & 61.4               &  77.0   & 79.8 & 50.5       \\ 
HG(ours)& ResNet50-IBN   & Occluded-Duke  & 65.1             &  \textbf{79.1}   & \textbf{81.4} & 54.7      \\ \hline 

\end{tabular}
}
    \caption{Accuracy of HG and state-of-the-art methods on the Occluded-Duke dataset.} 
\label{Tab:occ_dukeR}
\end{table}
\begin{table}[]
\centering
\resizebox{.7\linewidth}{!}{
\begin{tabular}{|l|l|l||rrr|r|}
\hline 
\textbf{Method}  & Backbone   &  \textbf{Supervision}  & \multicolumn{4}{c|}{\textbf{Accuracy}} \\ 
                    &   &                     & Rank-1  & Rank-5 & Rank-10 & mAP               \\\hline\hline    
IDE~\cite{zheng2015scalable}, ICCV 2015& -   & None   & 52.6    & 68.7   & 76.6    & 46.4                 \\
MLFN~\cite{chang2018multi}, CVPR 2018 & Custom    & None       & 42.3    & 60.6   & 68.5    & 36.0              \\
HACNN~\cite{li2018harmonious},CVPR 2018 & Custom         & None   & 29.1    & 44.7   & 54.7    & 26.1                       \\
PCB~\cite{sun2018beyond}, ECCV 2018 & ResNet50      & None     & 59.3    & 75.2   & 83.2    & 53.2                    \\
Part Bilinear~\cite{suh2018part}, ECCV 2018 & GoogLeNet & None & 54.9    & 70.8   & 77.7    & 50.3                         \\ \hline
teacher-S~\cite{teacherstudent}, ArXiv 2019 & ResNet  & Split Test Set  & 55.0    & 64.5   & 77.3    & 59.8           \\
PGFA~\cite{occ_pgfa}, ICCV 2019 & ResNet50         & PM & 57.1    & 77.9   & 84.0    & 56.2                          \\
PVPM~\cite{gao2020pose}, CVPR 2020 & ResNet50     & PM      & 70.4    & 84.1   & 89.8    & 61.2                     \\
HOReID~\cite{higherorder},  CVPR 2020 & ResNet50        & KP & 80.3    & -      & -       & 70.2                             \\ 
PAT~\cite{diverse} & ResNet50   & -  &  81.6    & -     &  -  & 72.1 \\ \hline \hline 
HG (ours Unsup)& ResNet50  & None  & 79.4 &  88.5  &  93.7 & 71.1                                           \\
HG (ours Sup) & ResNet50   & Occluded-Duke  & 82.3    & 89.7      &   94.1  & 71.7 \\ 
HG (ours Sup) & ResNet50-IBN   & Occluded-Duke  & \textbf{82.8}    & \textbf{90.1}     &  \textbf{ 94.6}  & 72.0 \\ \hline
\end{tabular}
}
    \caption{Accuracy of HG and state-of-the-art methods on the Occluded-ReID dataset. } 
\label{Tab:occ_reidR}
\vspace{-0.5cm}
\end{table}

\noindent \textbf{Results with Occluded and Partial ReID Problems.} 
Tabs.~\ref{Tab:occ_dukeR} and ~\ref{Tab:occ_reidR} show the result of our method on the Occluded-Duke and Occluded-ReID dataset compared with State-Of-The art (SOA) methods. ~\cite{occ_pgfa,depthocc,higherorder} are Occluded-ReID methods. We show the results for our method with both ResNet50 and ResNet-IBN~\cite{resnetIBN} backbones. In the Table "PM"-pose maps, "KP"-key -point detection. Our method outperforms all the other Occluded Person ReID methods mentioned in the table.  Our proposed HG method performs competitively in Rank-1 without any external input, such as pose or segmentation masks, on the Occluded-Duke dataset and 2.5\% on the Occluded-ReID dataset. Since the Occluded-ReID dataset does not contain training images, we show two sets of results, one with student trained on artificially occluded Market1501 dataset -HG(Unsup) and the other the student model trained on occluded samples from Occluded-Duke dataset -HG(Sup). Our results show that student trained on augmented Market1501 already outperforms many SOTA. Additionally, when the student is trained on a totally different occluded dataset (Occluded-Duke) from that of the test dataset (Occluded-ReID) still outperforms all SOTA. We also evaluate our method on partial ReID datasets~\cite{partialdata} and the result table is presented in the supplementary material. We can observe from the results that our method has improved over the SOTA by 2\%.

\begin{table}[!b]
\vspace{-0.4cm}
\centering
\resizebox{.75\linewidth}{!}{
\begin{tabular}{|l|l|l|r|}
\hline
\textbf{Model}   & \textbf{Experiment}    & \textbf{Training Data (Student)}  & \textbf{Rank-1}  \\    \hline \hline\multicolumn{4}{|l|}{\textbf{Impact of Training Architecture (Test Set: Occluded-ReID)}} \\ \hline 
\begin{tabular}[c]{@{}l@{}}HG (student+teacher)\end{tabular} & w/o occlusion classifier      & Augment Holistic & 75.6   \\
 & w/ occlusion classifier  & Augment Holistic  & 79.4                                                                       \\
 & w/o occlusion classifier& Occluded-Duke  & \textbf{82.3}                                                                  \\ \hline \hline
\multicolumn{4}{|l|}{\textbf{Impact of Losses (Test Set: Occluded-ReID)}} \\  \hline
teacher only      & identity only                 & -               & 59.5 \\
teacher+student & identity only              & Occluded-Duke    & 63.8  \\
teacher+student & identity+reconstruction                  & Occluded-Duke    & 67.7 \\
teacher+student & identity+reconstruction+MMD            & Occluded-Duke    & 76.1 \\ \hline 
teacher+student & MMD + attention (no autoencoder)                & Occluded-Duke    & 80.1  \\ 
teacher+student & reconstruction+MMD+attention & Occluded-Duke    & \textbf{82.3} \\ \hline \hline
\multicolumn{4}{|l|}{\textbf{Impact of backbone (Only ResNet50 Backbone, w/o Part Pooling)}} \\ \hline
teacher only & identity & -    & 51.3 \\ 
teacher+student & reconstruction+MMD+attention& Occluded-Duke    & 68.5 \\  \hline
\end{tabular}
}
    \caption{Impact on HG accuracy of training data and architecture on Occluded-ReID dataset.} 
\label{Tab:ablationS}
\end{table}

\begin{table}[]
\centering
\resizebox{.6\linewidth}{!}{
\begin{tabular}{|l|l|cc|cc|} 
\hline
 \textbf{Model} &\textbf{Backbone} & \multicolumn{2}{c|}{\textbf{Market1501}}  & \multicolumn{2}{c|}{\textbf{Duke-MTMC}} \\ 
 &  & Rank-1 &  mAP &  Rank-1 & mAP      \\
\hline \hline
PCB~\cite{sun2018beyond}, ECCV 2018 &  ResNet50                  & 92.3                       & 77.4                    & 81.8                           & 66.1                          \\
MaskReID~\cite{song2018mask}, CVPR 2018 & MSCAN                   & 90.0                         & 75.3                    & -                              & -                             \\
FPR~\cite{partial2}, ICCV 2019 & ResNet50                          & 95.4                       & \textbf{86.6}                    & 88.6                           & \textbf{78.4}                          \\
PVPM~\cite{gao2020pose}, CVPR 2020 & ResNet50                         & 93.0                         & 80.8                    & 83.6                           & 72.6                          \\
PGFA~\cite{occ_pgfa}, ICCV 2019 & ResNet50                        & 91.2                       & 76.8                    & 82.6                           & 65.5                          \\
HOReID~\cite{higherorder}, CVPR 2020 & ResNet50                     & 94.2                       & 84.9                    & 86.9                           & 75.6                          \\ \hline \hline 
MOS~\cite{aaaiOccluded}, CVPR 2021 & ResNet50-IBN                     & 95.4                      & 89.0                   & \textbf{90.6} & \textbf{80.2}   \\ 
PAT~\cite{diverse}, CVPR 2021 & ResNet50                     & 95.4                      & 88.0                   & 88.8 & 78.2   \\   
HG (Ours) & ResNet50                     & \textbf{95.6}                      & 86.1                   & 87.1 & 77.5   \\  \hline

\end{tabular}
}
\label{Tab:holisticR}
\caption{Accuracy of our HG method on Market1501 and Duke-MTMC datasets.} 
\vspace{-0.5cm}
\end{table}

\noindent \textbf{Ablation Study.}
An extensive study was conducted on Occluded-ReID to analyze the impact on training data and architecture performance. Results are shown in Tab.~\ref{Tab:ablationS}.

\noindent \textbf{(a) Impact of Data}. This study was performed on the full student-teacher model with all the components. In Tab.~\ref{Tab:ablationS}, "Augm. on Holistic": refers to student model trained with artificially occluded Market1501 dataset by Random Erasing~\cite{randomerasing}. We also use additional "occluded or non-occluded" binary classification on the features and show results with and without this classification. We can see that augmentation has helped the student model to learn Occluded-ReID and perform better than many SOTA. Using the Occluded-Duke to train the student shows that the student model performs even better on the Occluded-ReID dataset. Also, Occluded-Duke and Occluded-ReID datasets have no overlap.

\noindent \textbf{(b) Impact of Architecture.} To analyze the effect of architecture on the result from above, the teacher alone (pre-trained on Market1501) is fine-tuned on Occluded-Duke. Results indicate that fine-tuning the baseline network on Occluded-Duke performs poorly on Occluded-ReID. In Tab~3, the remarks "identity+reconstruction" refer to the experiment where  student-teacher model has been trained with identity loss and reconstruction loss alone. The reconstruction loss has helped the student to learn some generative properties, and hence the overall result on Occluded-ReID data is better than the baseline fine-tuned on the Occluded-Duke dataset. We further extend our experiments by using distribution loss, identity loss along with reconstruction loss, but no attention "reconstruction+MMD."  The "MMD+attention" experiment does not use reconstruction loss, and hence the autoencoder is not trained. Finally, we train the student model with the full system "reconstruction+MMD+attention," which includes the attention mechanism learned by matching distributions. 

\noindent \textbf{(c) Impact of CNN Backbone.} In order to assess the impact of part pooling, we perform another the "Impact of backbone" experiment, where only a ResNet-50 backbone is used with no part pooling. Although the overall results are this case is lower than when using part pooled features, it can be seen that proposed loss has still improves Occluded-ReID performance. From the results, we can conclude that our system learns occluded-ReID by using either artificially-occluded examples or real-occluded examples with holistic data as a reference. We show additional ablation studies in the supplementary materials.

\noindent \textbf{(d) Holistic ReID Datasets.} Tab.~\ref{Tab:holisticR}  shows results with the HG model on holistic Market1501~\cite{zheng2015scalable} and Duke-MTMC-\cite{dukeDUPLICATED} datasets.  Results show that HG is competitive with other SOTA models designed to work with the Occluded-ReID problem. 
\begin{figure}[h!]
 \centering
\includegraphics[width=.7\linewidth]{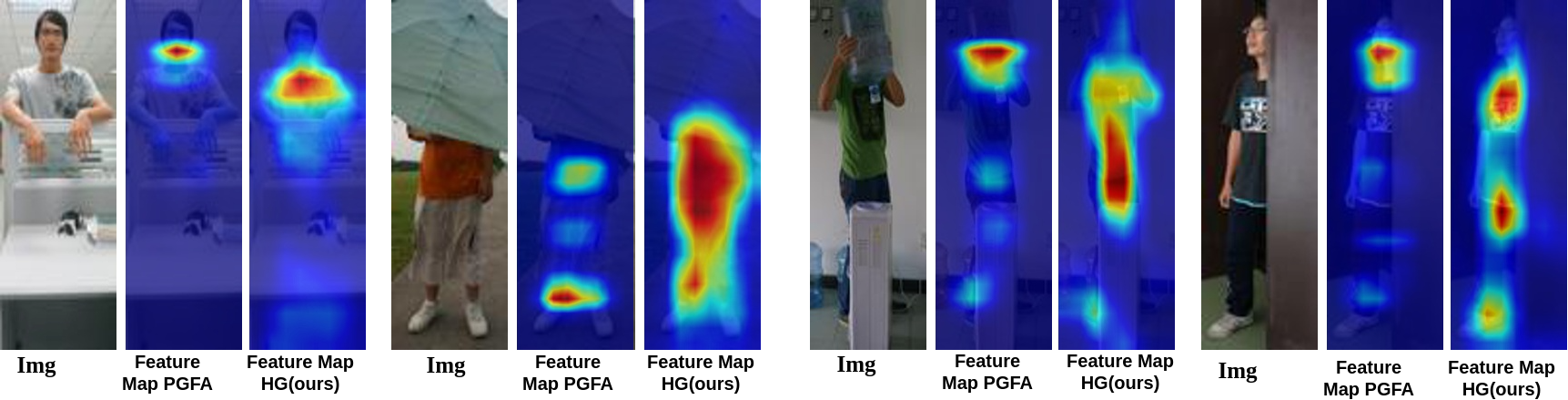}
   \caption{Activation maps generated for four occluded images of the Partial ReID dataset by the PGFA method~\cite{occ_pgfa} versus our HG method. PGFA uses pose estimation additionally.}
  \label{fig:heatmaps_ae_embed} 
  \vspace{-0.35cm}
\end{figure}

\noindent \textbf{Qualitative Results.} Fig.~\ref{fig:heatmaps_ae_embed} compares activation maps from the student model tested on examples with Partial-ReID dataset with activation maps of~\cite{occ_pgfa} (uses pose maps for attention). In the figure, both head and legs are occluded. But, from the activation maps of our proposed HG method, it can be seen that our method is good at localising non-occluded regions alone.

\vspace{-0.2cm}
\section{Conclusion}
\vspace{-0.2cm}
In this paper, a novel HG student-teacher model is proposed for occluded person ReID that only requires image identity labels but no costly process to focus on visible parts of occluded regions. The proposed HG teacher considers the DCD of among samples in a holistic dataset to train a student model to generate attention maps, thereby alleviating the occlusion problem. Unlike most methods in the literature that use external supervision (such as pose) to generate visibility cues, we only rely upon the distribution of holistic data during training, using it as a soft label. Hence during test time, our model requires no external cues such as pose, and the overall parameters include just the backbone Encoder and a small embedding to generate attention maps to be used during feature extraction. Joint learning of a denoising autoencoder was used to improve the ability to self-recover from occlusion. Results on several challenging datasets show that our HG method can outperform state-of-the-art models for Occluded-ReID, as well as Holistic ReID tasks.

\noindent \textbf{Acknowledgments:} This research was supported by the Compute Canada and MITACS.

\FloatBarrier

\newpage
\bibliography{latex/biblio}
\end{document}